\documentclass[letterpaper, 10 pt, conference]{ieeeconf}  

\IEEEoverridecommandlockouts                              

\usepackage{graphicx}
\usepackage{amsmath}
\usepackage{enumerate}
\usepackage{algorithm}
\usepackage{algorithmic}
\usepackage{cite}
\makeatletter
\let\NAT@parse\undefined
\makeatother
\usepackage[colorlinks,linkcolor=blue, anchorcolor=blue, citecolor=blue]{hyperref}
\usepackage{amssymb}
\usepackage{enumerate}
\usepackage{mathrsfs}
\usepackage[capitalize]{cleveref}
\usepackage{threeparttable}
\usepackage{bm}
\usepackage{float}
\usepackage{booktabs}
\usepackage{multirow}
\usepackage{xspace}
\usepackage{subfig} 
\usepackage{subfloat}
\newcommand{\ie}{{\emph{i.e.}}\xspace}
\newcommand{\eg}{{\emph{e.g.}}\xspace}

\newcommand{\etal}{{\emph{et al.}}\xspace}
\newcommand{\etc}{{\emph{etc.}}\xspace}

\title{\LARGE \bf
DisGNet: A Distance Graph Neural Network for Forward Kinematics Learning of Gough-Stewart Platform}

\author{Huizhi Zhu$^{1}$, Wenxia Xu$^{1,*}$, Jian Huang$^{2}$,~\IEEEmembership{Senior Member,~IEEE}, Jiaxin Li$^{1}$
\thanks{This work was supported by National Natural Science Foundation Youth Fund of China under Grant 61803286, and Innovation Fund Project of Hubei Key Laboratory of Intelligent Robot under Grant HBIRL202210.}
\thanks{$^{1}$Huizhi Zhu, Wenxia Xu and Jiaxin Li are with the Hubei Key Laboratory of Intelligent Robot, Wuhan Institute of Technology, Wuhan 430205, China (zhuizhi033@gmail.com; xuwenxia@wit.edu.cn; lijiaxin232@gmail.com).}
\thanks{$^{2}$Jian Huang is with Key Laboratory of Image Processing and Intelligent Control, School of Artificial Intelligence and Automation, Huazhong University of Science and Technology, Wuhan 430074, China (huang\_jan@mail.hust.edu.cn). }
\thanks{ $^{*}$\emph{ Corresponding author : Wenxia Xu}}}
\begin{document}

\maketitle
\thispagestyle{empty}
\pagestyle{empty}

\begin{abstract}
In this paper, we propose a graph neural network, DisGNet, for learning the graph distance matrix to address the forward kinematics problem of the Gough-Stewart platform. DisGNet employs the k-FWL algorithm for message-passing, providing high expressiveness with a small parameter count, making it suitable for practical deployment. Additionally, we introduce the GPU-friendly Newton-Raphson method, an efficient parallelized optimization method executed on the GPU to refine DisGNet's output poses, achieving ultra-high-precision pose. This novel two-stage approach delivers ultra-high precision output while meeting real-time requirements. Our results indicate that on our dataset, DisGNet can achieves error accuracys below 1mm and 1deg at 79.8\% and 98.2\%, respectively. As executed on a GPU, our two-stage method can ensure the requirement for real-time computation. Codes are released at \url{https://github.com/FLAMEZZ5201/DisGNet}.

\end{abstract}

\section{Introduction}

One important concern in robotics is the mapping relationship between a joint space and the pose space of its end-effector, which is commonly referred to as the kinematics problem. The solution of the kinematics problem is integral to both motion planning and control in robotics. 
Notably, the kinematic analysis of robots with complex or closed-loop structures can be a challenging task. The Gough-Stewart platform (GSP)~\cite{stewart1965platform} is one of the classic parallel mechanisms that features a closed-loop design. As a result of the distinct characteristics of parallel robot structures, determining its forward kinematics problem can be exceptionally complicated~\cite{dasgupta2000stewart}. 
While various works have been proposed to address this problem, there is currently no universally acknowledged method that can guarantee both the precision and efficiency of the solution simultaneously. There is still a strong desire in the parallel manipulator community to explore novel and stronger forward kinematics methods.

Neural networks (NNs) are a well recognized method for learning the mapping relationship and performing real-time solution. For forward kinematics learning in parallel manipulator, this started by via a single layer of multi-layer perceptrons (MLPs)~\cite{geng1991neural, seng1997forward}. In the study of non-parallel structures~\cite{grassmann2022dataset,grassmann2019merits}, NNs are also employed to solve forward kinematics. Despite NNs have cracked up new avenues for addressing complex mechanisms' forward kinematics, the majority of NNs in employ are intuitive MLPs with Euler angles as the output representation. On limited dataset, achieving high-precision pose outputs is difficult with this traditional learning method. The primary reason is that traditional MLPs have weak inductive biases~\cite{battaglia2018relational} on the one hand.  Inductive biases in deep neural networks are vital to many advances. Directly regressing rotation, on the other hand, requires the finding of better differentiable non-Euclidean representations, \ie, the rotation representation of the NN must be beyond four dimensions~\cite{zhou2019continuity}.

Motivated by exploring a network that can provide both  high-precision pose outputs and real-time solutions. We attempt to take advantage of the complex structure of parallel manipulators to build a graph and apply  message-passing graph neural network (GNN) to learn. Since its permutation equivariance, GNN has been successfully applied in numerous of fields~\cite{stokes2020deep}. Applying GNN techniques, however, raises an important question: for forward kinematics learning, what should be the input representation for GNN? GNNs typically use node coordinates as input features, but in forward kinematics, the robot's coordinate information are not known in advance. In the case of GSP, the physical connection lengths are more widely known. To overcome the challenge of unknown coordinate information, we opt for the graph distance matrix as the input for GNN in this work. 
\begin{figure}[t]
\centering
\includegraphics[width=8.6cm]{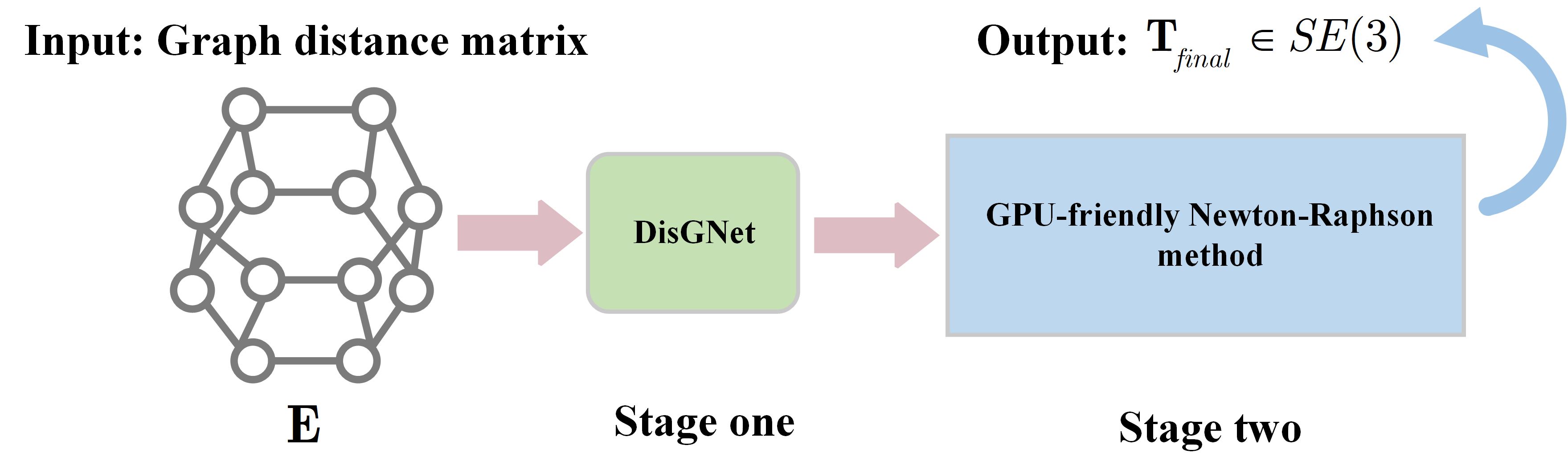}
\caption{The pipeline of our proposed forward kinematics solver for the Gough-Stewart platform.} 
\label{fig:1}
\end{figure}

Recently, several studies have shown that traditional GNNs struggle to use the distance matrix adequately to learn geometric information~\cite{zhang2021physically, schutt2021equivariant,pozdnyakov2022incompleteness}. This limitation follows primarily from traditional GNNs' limited expressive power~\cite{pozdnyakov2022incompleteness}. To explore better ways of utilizing the distance matrix as input for learning, li~\etal~\cite{li2023is} employed the k-WL/FWL algorithm to construct k-DisGNNs and achieved promising results. Inspired by this work, we use the k-FWL algorithm~\cite{cai1992optimal} to construct a high expressive power \textbf{Dis}tance \textbf{G}raph Neural \textbf{Net}work (DisGNet) for forward kinematics learning. The rotation in DisGNet is represented through 9D-SVD orthogonalization~\cite{levinson2020analysis}. DisGNet possesses a similar number of parameters as traditional MLP methods but achieves high-precision pose outputs.

To achieve higher-precision poses, it is standard to use the Newton-Raphson method to optimize the estimated pose, forming a two-stage~\cite{parikh2005hybrid} (NN+Newton-Raphson). However, unlike NN, the iterative computations of the Newton-Raphson method are time-consuming when large-scale poses are needed to optimize and do not support efficient parallel computation on GPU. To deal with this, we extend the Newton-Raphson method to GPU platform, presenting a novel two-stage framework (DisGNet+GPU-friendly Newton-Raphson), as shown in~\cref{fig:1}. The GPU-friendly Newton-Raphson achieves GPU acceleration by efficiently approximating the Moore-Penrose inverse using matrix-matrix multiplication to enable real-time computations on GPU platform.

Our contributions are summarized as follows:
\begin{itemize}
\item We propose a highly expressive power GNN, namely DisGNet, which is strong at learning geometric features represented by distances and provides high-precision poses. DisGNet has fewer parameters while producing results comparable to models with more parameters.
\item We propose a GPU-friendly Newton-Raphson method, a two-stage framework that extends Newton-Raphson to the GPU for GPU computation. This novel two-stage framework ensures ultra-high precision output and forward kinematics solving in real-time.
\item We propose refined metrics to evaluate the performance of forward kinematics learning models. Moreover, we provide the dataset and  benchmark results, which will benefit to the development of learning-based kinematics methods for parallel manipulator community. 
\end{itemize}
\section{Related Work}
\textbf{Forward kinematics methods of GSP. }The Gough-Stewart platform, published by Stewart\cite{stewart1965platform} in 1965, introduced the structure of this mechanism, making the forward kinematics problem of parallel mechanisms a hot topic in this field. Despite numerous works attempting to address this problem, it has never been completely resolved up to now. Generally, we can categorize forward kinematics methods into the following four types:
\begin{itemize}
\item Analytical method.
\item Numerical method.
\item Sensor-based solving method.
\item Learning-based method.
\end{itemize}

The primary purpose of the \textbf{analytical method} is to derive a high-order equation with one unknown variable by through elimination. All possible platform poses are subsequently given by solving this equation. Among the mathematical techniques for equation elimination employed in the past are: interval analysis~\cite{merlet2004solving}, Gr{\"o}basis approach~\cite{dhingra1998grobner}, continuation method~\cite{raghavan1993stewart},~\etc  The analytical method often yields multiple closed-form solutions when encountering complex structures, \eg, 6-6 GSP structure (where $m$-$n$ specifies the number of hinge points on the base and moving platforms), and this method is only applicable to specific configurations. Besides, the analytical method is typically computationally inefficient and unsuitable to deploy in real-time work systems\cite{dasgupta2000stewart}. For practicality, some researchers are opt for \textbf{numerical method} to solve this problem~\cite{265928,tarokh2007real}. The goal of the numerical method is to build the objective function via the GSP's inverse kinematics problem, afterwards to leverage algorithms for non-linear optimization to determine the objective value iteratively. One prominent numerical method for solving the kinematics problem with parallel robots is the Newton-Raphson  algorithm\cite{ypma1995historical}. It's a first-order Taylor series-based non-linear optimization method. Unfortunately, the initial values selected for this iterative method are extremely vital, as choosing improper initial values can lead to divergence and lower the overall success rate. The \textbf{sensor-based solving method} is comparatively simpler and more practical among the other three methods\cite{schulz2018performance,merlet1993closed}. However, its applicability is limited due to assembly errors and suitability for only one specific configuration, rendering this method non-universal. The \textbf{learning-based method} in previous works has involved technologies such as support vector machines \cite{morell2013solving} and NN. NN is becoming the prevalent way for this kind of work. Normally, a MLP serves as the model to learn the end-effector pose (rotation representation is performed via Euler angles), and the labeled ground truth is used to minimize the mean squared error (MSE). We divide  NN-based methods into two categories in this work. One adopts neural networks for \textbf{direct regression way}\cite{geng1991neural, seng1997forward}, yielding end-to-end output results. The other adopts a \textbf{two-stage way}\cite{parikh2005hybrid, he2024phynrnet}. Some researchers utilize Newton-Raphson after network output to improve the result in order to get better precision because MLP network regression often yields poorer precision. This theoretically overcomes the issue of lower MLP's output precision while providing initial values for Newton-Raphson. In fact, it tends to be challenging to learn poses within a wide variety of motion spaces when dependent only on MLP with limited inductive biases. Learning via physics-informed neural networks \cite{he2024phynrnet,merlet2023advances} has been the focus of some researchers recently. \textbf{In our work, we focus a two-stage way to explore a GNN with high expressiveness that can support end-to-end  high-precision pose outputs and guarantee that its inference time satisfies practical operational requirements}. 

\textbf{Message-passing GNNs and its expressiveness. } The majority of GNNs~\cite{kipf2016semi} are made up of two parts. Primarily an underlying message-passing mechanism that aggregates the local 1-hop neighborhood data to compute node representations. The following is an $L$-layer stack that combines $L$-hop neighborhood nodes to improve network expressivity and facilitate information transfer between $L$-hops apart. Xu \etal~\cite{xu2018powerful} showed that the Weisfeiler-Leman (WL) test~\cite{weisfeiler1968reduction} limits the expressiveness of message-passing GNNs, \ie,  the expressiveness of traditional message-passing GNNs$\leq$1-WL.  It has been demonstrated that the expressiveness upper-bounded by 1-WL is insufficient for distinguishing non-isomorphic graphs or to capture the distance information from weighted undirected graph edges~\cite{pozdnyakov2022incompleteness}. It was suggested to employ high-order GNNs based on k-WL to improve the its expressiveness~\cite{morris2020weisfeiler}. The Folklore WL (FWL) test~\cite{cai1992optimal} was used by Maron \etal~\cite{maron2019provably} to build GNNs, and Azizian \etal~\cite{azizian2020expressive} demonstrated that these GNNs are the most potent GNNs for a particular tensor order. Li \etal~\cite{li2023is} proposeed k-DisGNNs by applying the k-WL/FWL test, which efficiently takes advantage of the rich geometry present in the distance matrix.

\section{Preliminaries}
\begin{figure}[H]
\centering
\subfloat[]{
\includegraphics[width=3.8cm]{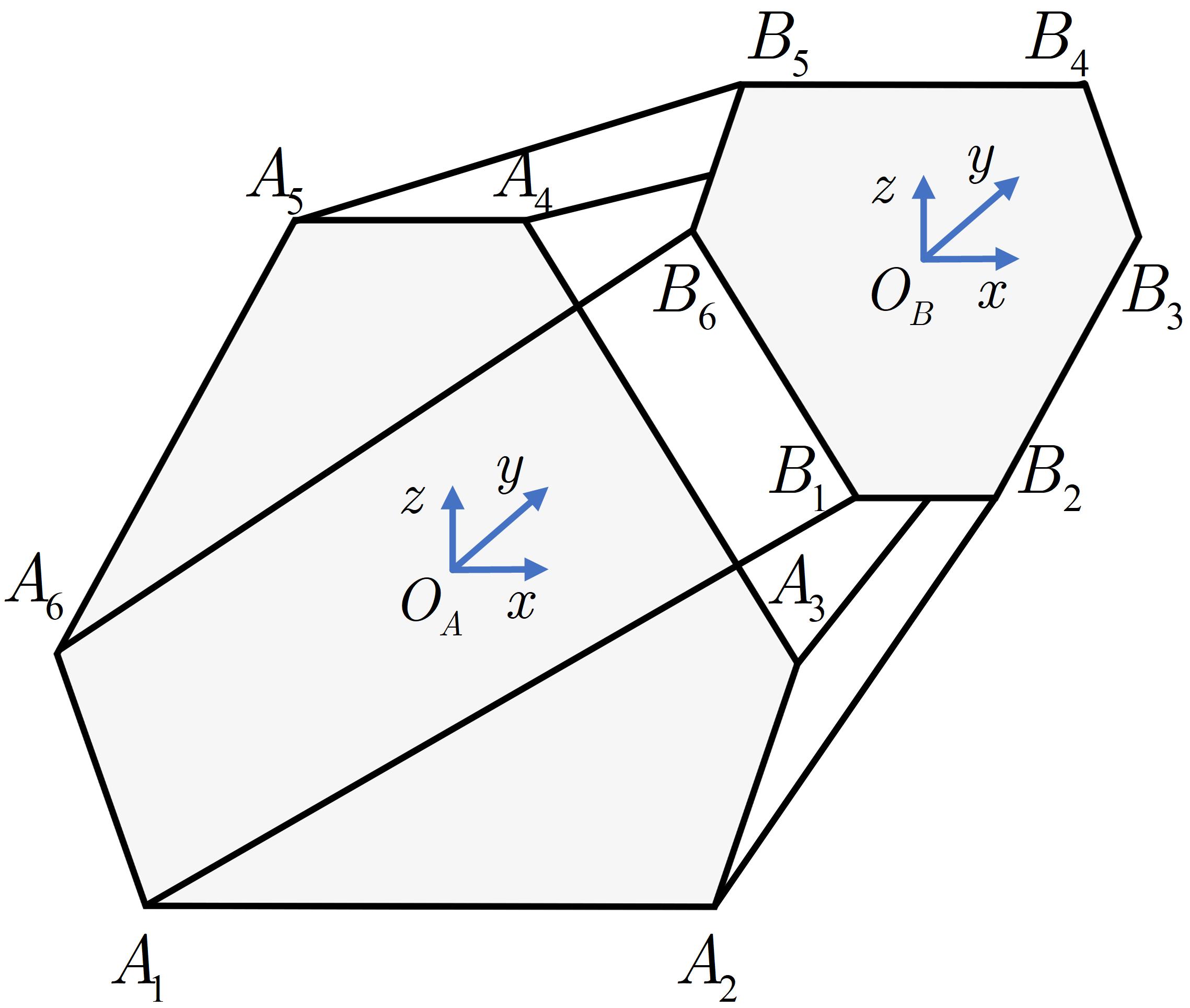}
}
\subfloat[]{
\includegraphics[width=3.7cm]{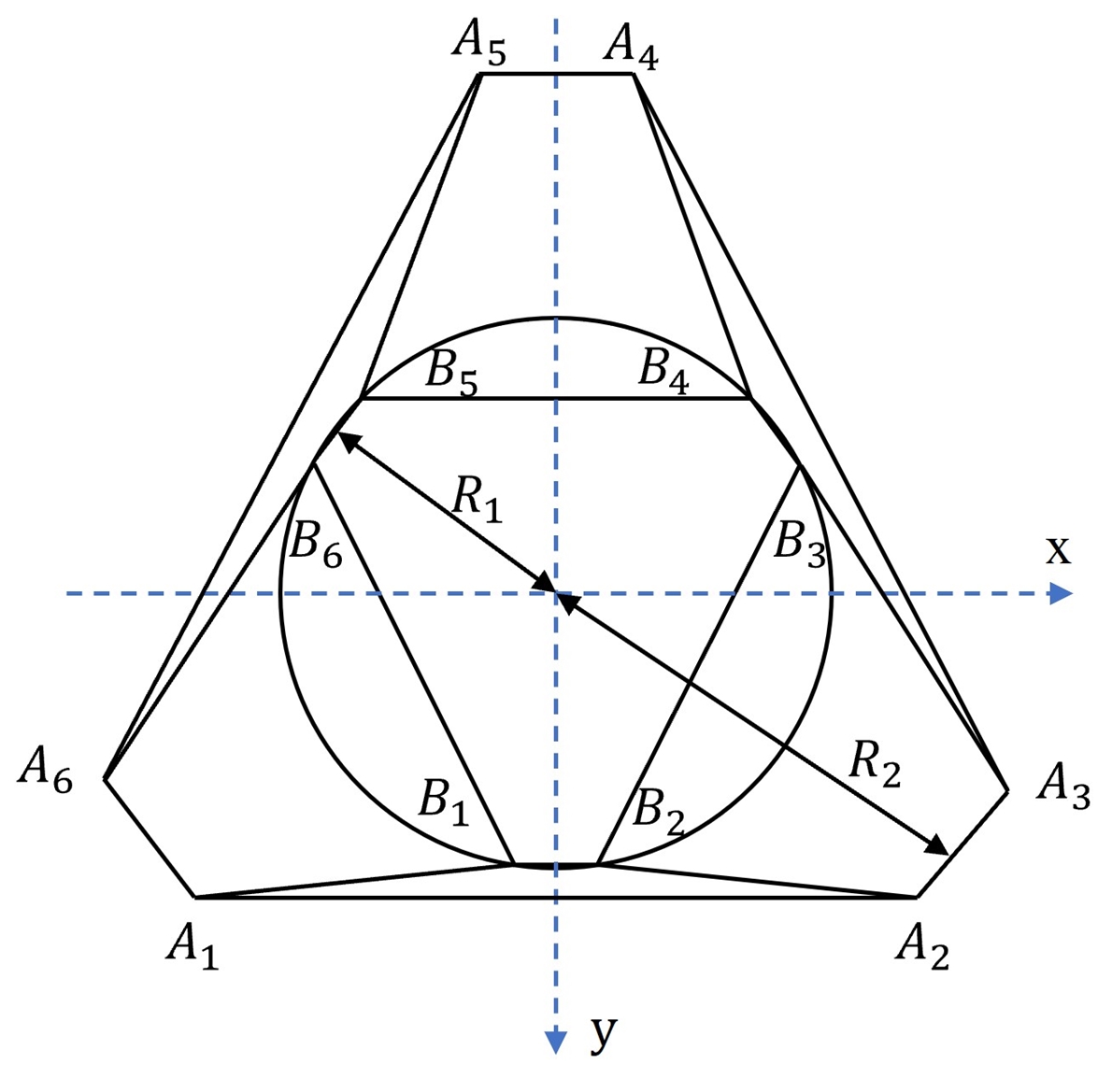}
}
\caption{Illustration of the 6-6 GSP structure. (a) Initial coordinate systems. (b) Perspective planar view.}
\label{fig:2}
\end{figure}
Before describing the problem and its definition, let's first provide a brief introduction to the 6-6 GSP (6-DoF parallel mechanism) structure and its coordinate relationships. Typically, a 6-6 GSP comprises a base platform, a moving platform (\textit{aka} the end-effector), and six displacement actuators that connect these two platforms. In~\cref{fig:2}, we offer the relevant structural diagram. We define the base coordinate system $\{A\}$ using the geometric center ${O_A}$ of the base platform, and establish the moving coordinate system $\{B\}$ using the geometric center ${O_B}$ of the moving platform. The hinges are denoted as $A_s$ and $B_s$ in the base and moving coordinate systems, respectively, while $\mathbf{l}_s$ represents the initial lengths of the actuators between these hinges, $s=1,\cdots,6$. $R_1$ and $R_2$ respectively refer to the radii of the base platform and the moving platform.

\subsection{Notions}
\newtheorem{problem}{\bf  Problem}[section]
\begin{problem}[Forward kinematics problem]
\label{p:3.1}
\textit{Given the robot's  joint information vector $\bm{\theta}\in\mathcal{C}$ to determine the robot's end-effector pose $\bm{\tau}\in\mathcal{T}$ is referred to as the forward kinematics problem, where $\mathcal{C}$ denotes the robot's configuration space, $\mathcal{C}\in\mathbb{R}^{n}$, and $\mathcal{T}$ represents the robot's end-effector workspace. When the end-effector has 6-DoF, it is commonly defined that its pose is represented as $\mathbf{T}=\bm{\tau}$, $\mathbf{T}\in SE(3)$. The forward kinematics mapping function is denoted as $\mathcal{FK}: \bm{\tau}\rightarrow\mathcal{T}$.}
\begin{equation}
\mathcal{FK}(\bm{\theta})=\bm{\tau}\in\mathcal{T}
\end{equation}
\end{problem}
\begin{problem}[Inverse kinematics problem]
\label{p:3.2}
\textit{Given the pose information of the robot's end-effector $\tau$ to determine the robot's joint vector $\bm{\theta}$ refers to the inverse kinematics problem. Typically, we use $\mathbf{T}$ to denote its 6-DoF end-effector pose information. The inverse kinematics mapping function is represented as $\mathcal{IK}:\mathcal{T}\rightarrow\bm{\tau}$.}
\begin{equation}
\mathcal{IK}(\mathbf{T})=\bm{\theta}\in\mathcal{C}
\end{equation}
\end{problem}
\newtheorem{Mydef}[problem]{\bf Definition}
\begin{Mydef}[Graph representation]
\label{d:3.3}
\textit{We define the 6-6 GSP structure as a graph $\mathcal{G}=(\mathcal{V}, \mathcal{E}, \mathbf{A})$, where $\mathcal{V}$ is the set of nodes (\ie consists of hinge points $A_s$ and $B_s$) and $\vert\mathcal{V}\vert = N$. The physical connections between nodes are represented by a set of edges called $\mathcal{E}$. $\mathbf{A}$ denotes the adjacency matrix of $\mathcal{G}$, $\mathbf{A}\in \mathbb{R}^{N\times N}$. In this work, we avoid redundant edges in the graph, \ie, the connections between hinge points of the same platform, resulting in the final graph being an octahedral structure, where $N=12$.}
\end{Mydef}

\begin{Mydef}[Graph distance matrix]
\label{d:3.4}
\textit{We define the distance weight $d_{ij}$ for each edge ${e_{ij}}$ in the GSP's graph, where ${e_{ij}} \in \mathcal{E}$. The graph distance matrix $\mathbf{E}$ is obtained through the function $\mathcal{P}$ by mapping the adjacency matrix $\mathbf{A}$ and the distance weights $d_{ij}$. Here, $\mathbf{E}\in\mathbb{R}^{N \times N}$. The mapping function $\mathcal{P}$ is described below:}
\begin{equation}
\mathbf{E}=\mathcal{P}(\mathbf{A},d_{ij})=\left\{ \begin{array}{ll}
a_{ij} = d_{ij},&if\ a_{ij} = 1;\\
a_{ij} = 0, &otherwise.\\
\end{array}\right.
\end{equation}
\textit{where $a_{ij}$ denotes the element in the $i$-th row and $j$-th column of the adjacency matrix $\mathbf{A}$.}
\end{Mydef}

\subsection{Inverse kinematics of Gough-Stewart platform}

In this subsection, we established the inverse kinematics model for the GSP. As depicted in \textbf{problem}~
\ref{p:3.2}, the inverse kinematics problem for the GSP refers to obtaining the joint displacement variables $\mathbf{l}_{os}$ given the moving platform pose $\mathbf{T}$. Here, $\mathbf{l} _{os}= \bar{\mathbf{l}}_{s}-\mathbf{l}_{s}$, and $\mathbf{l}_{os} \in \mathbb{R}^{6}$.  $\bar{\mathbf{l}}_{s}$ represents the length of the actuators in the current state, while $\mathbf{l}_{s}$ refers to the length of the actuators in the initial state. The inverse kinematics mapping function $\mathcal{IK}: \mathbf{T}\rightarrow\mathbf{l}_{os}$ for the GSP is as follows:
\begin{equation}
\label{eq:4}
\mathbf{l} _{os} = \mathcal{IK}(\mathbf{T})= \bar{\mathbf{l}}_{s}-\mathbf{l}_{s} = \Vert\mathbf{T}\mathbf{b}_s-\mathbf{a}_s\Vert_2-\mathbf{l}_{s}
\end{equation}
where $\mathbf{T}= \begin{pmatrix}\mathbf{R}&\mathbf{t}\\
0&1\end{pmatrix}_{4\times4}$ is a homogeneous transformation matrix, $\mathbf{t}= (x\ y\ z)\in\mathbb{R}^{3}$ is a translation variable, $\mathbf{R}=\{\text{exp}(\omega^{\wedge})\in SO(3)\vert \text{det}(\mathbf{R}\mathbf{R}^{\top})=1\}$ is a rotation matrix, ${\omega}^{\wedge}$ is skew-symmetric matrix, the mapping operator $^{\wedge}:{\omega}\in\mathbb{R}^{3}\to{\omega}^{\wedge}\in\mathfrak{so}(3)$, and  $\text{exp}(\cdot)$ denotes exponential mapping of Lie group.
$\mathbf{a}_s = (A_{sx}\ A_{sy}\ A_{sz}\ 1)^{\top}$, $\mathbf{b}_s = (B_{sx}\ B_{sy}\ B_{sz}\ 1)^{\top}$ are constant vectors of the lower and upper hinge points in the coordinate system $\{A\}$ and $\{B\}$ respectively. 

\subsection{Problem statement}
As depicted in \textbf{problem}~\ref{p:3.1}, the forward kinematics problem for the GSP refers to obtaining the moving platform pose $\mathbf{T}$ given the joint displacement variables $\mathbf{l}_{os}$. In previous work, Euler angles were commonly used to represent rotations, with the output pose being denoted as $(x, y, z, \alpha, \beta, \gamma)$. These pose were learned using a MLP, which served as its forward kinematics mapping function, expressed as $MLP: \mathbb{R}^{6} \rightarrow \mathbb{R}^{6}$ . 
\begin{figure*}[t]
\centering
\includegraphics[width=16.5cm]{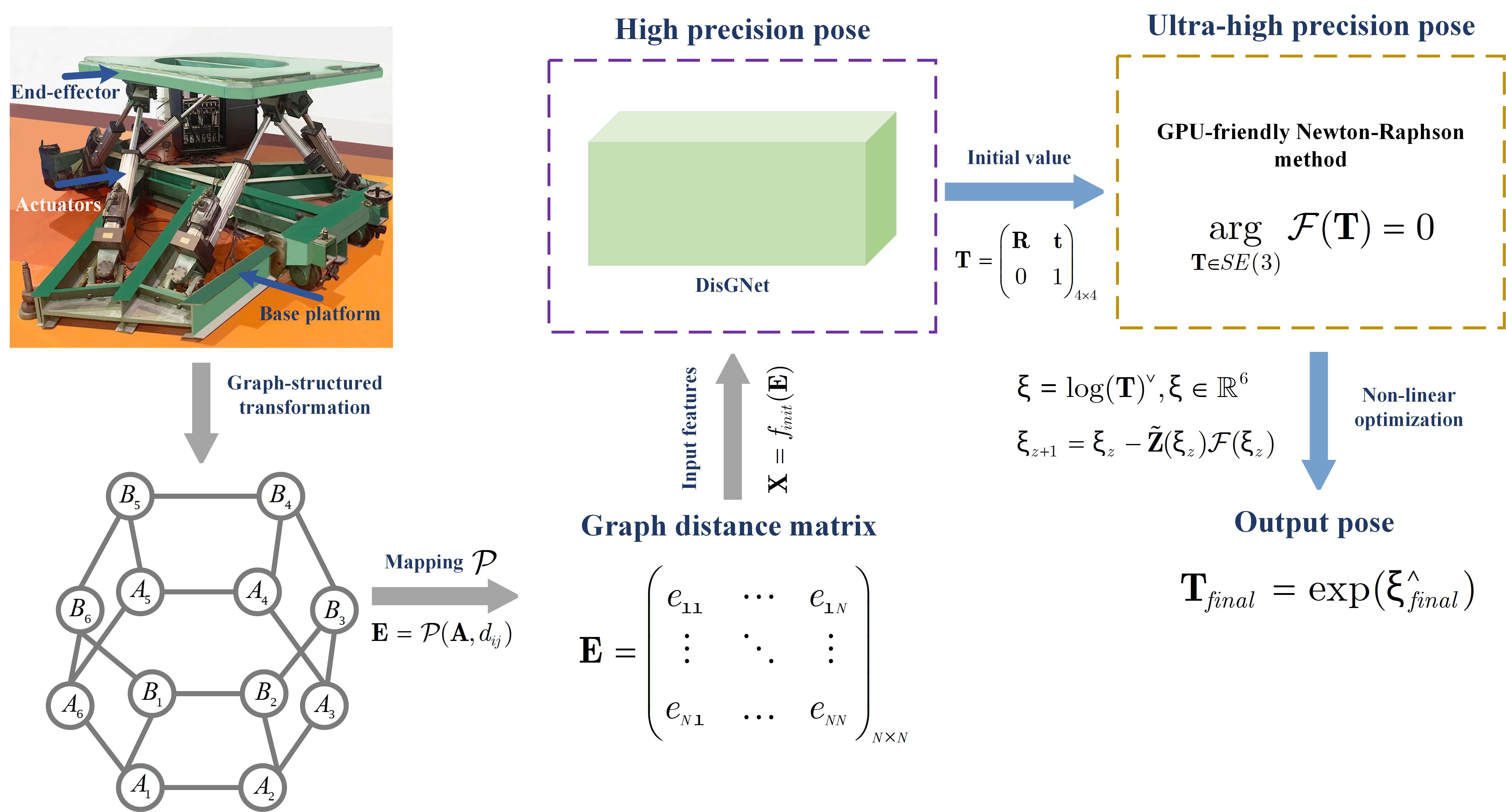}
\caption{Our forward kinematics solver for the Gough-Stewart platform is based on a two-stage framework. DisGNet learns the geometric information from the distance representation of graph and outputs end-effector pose $\mathbf{T}$ with high precision in an end-to-end manner. Subsequently, with the estimated value $\mathbf{T}$ as the initial value, the Newton-Raphson method is utilized for optimization. It offers ultra-high precision for the end-effector pose and can optimize fast via GPU parallel processing.} 
\label{fig:3}
\end{figure*}
In this study, we construct the graph of GSP and utilize its graph distance matrix $f_{init}(\mathbf{E})$ as the input feature $\mathbf{X}$ for learning the end-effector pose $\mathbf{T}$ through DisGNet. The forward kinematics mapping function can be briefly formulated as $\mathbf{T} = \mathcal{G}(\mathbf{X})$. The objective is to find the model parameters which minimize the error
between the ground truth pose $\mathbf{T}_{gt}$ and estimated pose $\mathbf{T}$:
\begin{equation}
\Theta^{\ast} = \arg\mathop{\min}\limits_{\Theta}\mathcal{L}(\mathcal{G}(\mathbf{X}),\mathbf{T}_{gt}, \Theta)
\end{equation}
where $\Theta$ is the set of parameters in the DisGNet model $\mathcal{G}$.

\section{Proposed Method}

\subsection{Overview}
In~\cref{fig:3}, we illustrates the two-stage framework of our proposed forward kinematics solver, which consists of the following three components:

\begin{itemize}
\item \textbf{Constructing the graph representation and distance matrix of GSP.}  Establishing learnable graph-structured data for the 6-6 GSP is vital before deploying DisGNet for forward kinematics learning. In this component, we overcome the challenge of unknown vertex coordinate information by implementing the known distance information of GSP as input representation.
\item \textbf{Stage~\uppercase\expandafter{\romannumeral1}: Utilizing DisGNet to attain high precision pose.} In this component, a DisGNet is proposed a to learn graph distance matrix. DisGNet is capable of high expressiveness while meeting the real-time inference requirements for learning graph distance matrices. The initialization block, message passing block, and prediction head are the three parts of our DisGNet. For the prediction head output, we build a weighted loss function, where SVD orthogonalization~\cite{levinson2020analysis} produces the rotation output $\mathbf{R}$.
\item \textbf{Stage~\uppercase\expandafter{\romannumeral2}: Employing the GPU-friendly Newton-Raphson method to quickly acquire ultra-high precision pose.} In this component, the GPU-friendly Newton-Raphson method is proposed to optimize the initial values provided by DisGNet. This particular Newton-Raphson method is compatible with GPU computation. We accelerate the process on GPU by efficiently approximating the Moore-Penrose inverse using matrix-matrix multiplication.
\end{itemize}

\subsection{Graph representation and distance matrix of GSP}
Before implementing DisGNet, the graph representation of GSP must be defined, and the graph-structured data is required for the network's input. According to the description provided in \textbf{definition}~\ref{d:3.3}, the nodes $\mathcal{V}$ of the graph represent the hinge points within the 6-6 GSP structure, while the edges $\mathcal{E}$ represent the physical connections. Thus, we have obtained a graph representing an octahedral structure, denoted as $\mathcal{G}=(\mathcal{V}, \mathcal{E}, \mathbf{A})$.

The 2D/3D coordinate information of nodes often serves as the main input feature in studies pertaining to GNNs. Nevertheless, it not attainable to obtain the node coordinate information in advance for the forward kinematics problem. Therefore, choosing which features to use as input presents the primary obstacle when building the network. The graph distance matrix is used as a learning feature in this work. Since in GSP the matching physical connections have known lengths, this suggests that the relevant nodes in the graph have known distances from each other. Following the description given in \textbf{definition}~\ref{d:3.4}, we can write the graph distance matrix for $\mathcal{G}$ as follows:
\begin{equation}
\mathbf{E}=
\begin{pmatrix}
e_{11} &\cdots&e_{1N}\\
\vdots &\ddots&\vdots\\
e_{N1} &\cdots&e_{NN}\\
\end{pmatrix}_{N\times N}
\end{equation}

\subsection{DisGNet}
\begin{figure}[t]
\centering
\includegraphics[width=8.6cm]{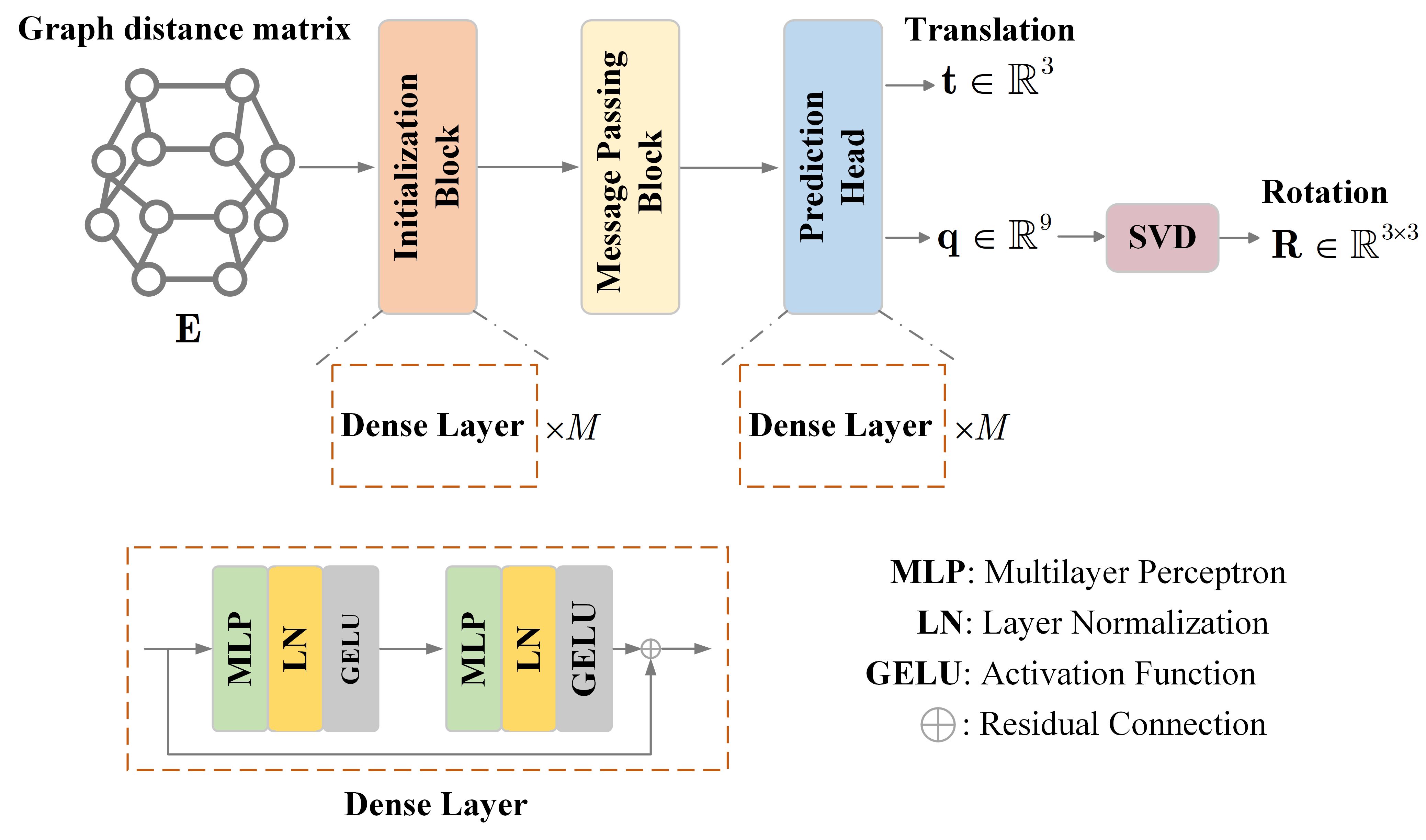}
\caption{The backbone of our proposed DisGNet.} 
\label{fig:4}
\end{figure}

In this subsection, we will introduce our proposed DisGNet. DisGNet can effectively leverage graph distance matrix for learning and enhance network representation capability by utilizing the 2-FWL algorithm, thereby yielding high-precision poses. Our DisGNet primarily consists of three components: the initialization block, message passing block, and prediction head, as depicted in ~\cref{fig:4}. We will introduce these three components in more detail below. 

\textbf{Initialization block.} DisGNet initializes  $k$-tuple $\bm{v}$ (in this work, $k=2$) with an injective function by:
\begin{align}
\mathbf{X}&=f_{init}(\mathbf{E})\notag\\ 
&= \mathop{\bigodot}\limits_{i\in[k]}f^{i}_{d}(f^{emb}_d(d_{v_i}))\odot\mathop{\bigodot}\limits_{i,j\in[k],i\neq j}f^{ij}_{e}(f^{rbf}_e(e_{v_iv_j}))
\end{align}
where $\odot$ denotes hadamard product, $[k]=1,\cdots,k$, $\bm{v}=(v_1,\cdots,v_k)\in\mathcal{V}^{k}$. $f^{emb}_d: \mathbb{Z}^{+}\rightarrow\mathbb{R}^{H_d}$ is embedding mapping function and  $f^{rbf}_e: \mathbb{R}^{+}\rightarrow\mathbb{R}^{H_e}$ is a radial basis function. $f^{i}_{d},f^{ij}_{e}$ are dense layer, which using MLPs that maps distance features and embeddings to the common feature space $\mathbb{R}^{H}$. The distance matrix $\mathbf{E}$ and the $d$ type of each node (one-hot format of node order) can be injectively embedded into the $k$-tuple by the initialization block.

\textbf{Message passing block.} Our message passing block enhances network's expressiveness through the k-FWL algorithm to learn the initial features $\mathbf{X}$. Here, we define the input features $\mathbf{X}$ as the tuple labels $h^{(0)}_{\bm{v}}$, and we specify the message passing function $f_{mp}:\mathbb{R}^{H} \rightarrow \mathbb{R}^{H}$:
\begin{equation}
h^{(t+1)}_{\bm{v}} = f_{mp}(h^{(t)}_{\bm{v}},\{\!\!\{\bm{H}^{(t)}_j(\bm{v})|j\in[N]\}\!\!\})
\end{equation}
\begin{equation}
\bm{H}^{t}_{j}(\bm{v})=(h^{(t)}_{\bm{w}}|\bm{w}\in G_j(\bm{v}))
\end{equation}
\begin{equation}
G_j(\bm{v})=\begin{pmatrix}(j,v_2,\cdots,v_k)&\cdots&(v_1,v_2,\cdots,j)
\end{pmatrix}
\end{equation}
where $\{\!\!\{\}\!\!\}$ represent mutiset. $G_j(\bm{v})$ is the set of neighbors of $k$-tuple $\bm{v}$. For more introduction and implementation details about k-WL/FWL, please see~\cite{li2023is, maron2019provably}.

\textbf{Prediction head.} Our prediction head is to linearly mapping the features from the message passing block into a 12-dimensional vectors using a dense layer. A 9D rotation representation $\mathbf{q}\in\mathbb{R}^{9}$ is utilized to learn rotation, while translation is denoted as $\mathbf{t}\in\mathbb{R}^{3}$. Subsequently, the rotation matrix $\mathbf{R}$ is obtained through orthogonalization using an SVD layer. The out mapping function $f_{out}: \mathbb{R}^{H}\rightarrow\mathbb{R}^{12}$ is:
\begin{equation}
\mathbf{t},\mathbf{q}=f_{out}(\{\!\!\{h^{(final)}_{\bm{v}}\}\!\!\})
\end{equation}
\begin{equation}
\mathbf{R} = \text{SVD}(\text{M}(\mathbf{q})) = \mathbf{U}\mathbf{\Sigma}^{\prime}\mathbf{V}^\top
\end{equation}
where $\text{SVD}(\cdot)$ is singular value decomposition orthogonalization layer. ${\text{M}(\mathbf{\cdot})}$ denotes the vector-to-matrix operator, ${\text{M}(\mathbf{q})}: \mathbf{q}\in\mathbb{R}^9\rightarrow\mathbf{M}\in\mathbb{R}^{3\times3}$. $\mathbf{\Sigma}^{\prime}$ is ${diag}(1,1,\det(\mathbf{U}\mathbf{V}^\top))$.

\textbf{Loss function.} We minimize the Euclidean distance between the estimated value and ground truth by constructing the $l_2$-norm of the vectors. By building a continuous, smooth, and injective loss in Euclidean space, we can learn the translation and rotation of the end-effector. The loss functions for translation and rotation are denoted as $\mathcal{L}_{\text{trans}}$ and $\mathcal{L}_{\text{rot}}$, respectively.
\begin{equation}
\mathcal{L}_{\text{trans}}=\Vert{\mathbf{t}-{\mathbf{t}}_{gt}}\Vert_{2}
\end{equation}
\begin{equation}
\mathcal{L}_{\text{rot}}=\Vert{\mathbf{R}-{\mathbf{R}_{gt}}}\Vert_{F}
\end{equation}
\begin{equation}
\mathcal{L} = \mathcal{L}_{\text{trans}}+\beta\mathcal{L}_{\text{rot}}
\end{equation}
where $\mathbf{t}_{gt}$ represents the ground truth, $\mathbf{t}$ is the estimated value, and $\Vert{\cdot}\Vert_{2}$ indicates the $l_{2}$-norm. The Frobenius norm is indicated by $\Vert{\cdot}\Vert_{F}$, the estimated value is $\mathbf{R}$, and the ground truth is $\mathbf{R}_{gt}$. The final weighted loss function is $\mathcal{L}$. $\beta$ is scale factor, which balances between $\mathcal{L}_{\text{trans}}$ and $\mathcal{L}_{\text{rot}}$.

\subsection{GPU-friendly Newton-Raphson method}
In this subsection, we will present how to employ the GPU-friendly Newton-Raphson method to iteratively optimize the estimated poses that have been obtained from DisGNet. Here, we extend the Newton-Raphson method to the GPU platform, taking advantage of powerful GPU to accelerate up parallel computing and improve the ability to achieve ultra-high precision poses. The Newton-Raphson method is described in below.

\textbf{Newton-Raphson method.} To optimize the estimated pose $\mathbf{T}$, considered one has to establish an objective function before deploying the Newton-Raphson method. For forward kinematics problem, we obtain the objective function via inverse kinematics problem:
\begin{equation}
\mathcal{F}(\mathbf{T}) = \mathcal{IK}(\mathbf{T})-\mathbf{l}_{os}=0
\end{equation}
where $\mathcal{IK}(\cdot)$ is the inverse kinematics mapping function, which is given in~\cref{eq:4}, and $\mathbf{l}_{os}$ is the joint displacement variable of GSP. 

According to the given objective function $\mathcal{F}(\cdot)$, we can present its first-order Taylor expansion as: $\mathcal{F}(\xi)=\mathcal{F}(\xi_{0})+\mathbf{J}(\xi_{0})(\xi-\xi_{0})$.                                                            For the sake of convenience in calculation and derivation, we opt to represent pose using the Lie algebra $\xi$, $\xi = \text{log}(\mathbf{T})^{\vee}$, where ${\xi}= (q_1\ q_2\ q_3\ q_4\ q_5\ q_6)$, ${\xi}\in\mathbb{R}^{6}$, $\text{log}(\mathbf{T})={\xi}^{\wedge}\in\mathfrak{se}(3)$, and $\text{log}(\cdot)$ represents logarithm mapping of Lie group, the mapping operator $^{\vee}:
{\xi}^{\wedge}\rightarrow{\xi}$. And $\mathbf{J}$ is the Jacobian matrix, which can be define as below:
\begin{equation}
\mathbf{J} = \begin{pmatrix}
\frac{\partial{\mathcal{F}_1}}{\partial{{q}_1}}&\cdots&\frac{\partial{\mathcal{F}_1}}{\partial{{q}_6}}\\
\vdots&\ddots&\vdots\\
\frac{\partial{\mathcal{F}_6}}{\partial{{q}_1}}&\cdots&\frac{\partial{\mathcal{F}_6}}{\partial{{q}_6}}\\
\end{pmatrix}_{6\times6}
\end{equation}

By setting $\mathcal{F}(\xi)$ = 0, we can derive its iterative equation:
\begin{equation}
{\xi}_{z+1} = {\xi}_{z}-\mathbf{J}^{-1}({\xi}_{z})\mathcal{F}({\xi}_{z})
\end{equation}
where $z$ is iterations, $z=0,1,\cdots, \infty$. It should be note that the inverse of the Jacobian matrix might not exist in some structures, such as redundant actuation. We usually realize the Moore-Penrose inverse $\mathbf{J}^{+}$ to replace  $\mathbf{J}^{-1}$:
\begin{equation}
{\xi}_{z+1} = {\xi}_{z}-\mathbf{J}^{+}({\xi}_{z})\mathcal{F}({\xi}_{z})
\end{equation}

\textbf{GPU-friendly Newton-Raphson method.} Efficient computation of the Moore-Penrose inverse $\mathbf{J}^{+}$ becomes a consideration when extending the Newton-Raphson algorithm to the GPU platform. While SVD method is generally applicable for computation, its inefficiency in obtaining the Moore-Penrose inverse on a GPU arises from its non-compliance with matrix-matrix multiplication rules in GPU computation.
We employ an iterative technique from Razavi et al.~\cite{razavi2014new} to approximate the Moore-Penrose inverse via helpful matrix-matrix multiplications with the aim of accelerating the computation. This technique has also been adopted by linear-complexity transformer studies~\cite{xiong2021nystromformer}.
\newtheorem{theorem}{\bf  Theorem}[section]

\begin{theorem}
\label{th:4.1}
\textit{For Jacobian matrix $\mathbf{J}\in\mathbb{R}^{6\times6}$, given the initial approximation ${\mathbf{Z}_{0}}$,  we have:
\begin{equation}
\label{eq: 20}
 \mathbf{Z}_{f+1}=\frac{1}{4}\mathbf{Z}_{f}(13\mathbf{I}-\mathbf{J}\mathbf{Z}_{f}(15\mathbf{I}-\mathbf{J}\mathbf{Z}_{f}(7\mathbf{I}-\mathbf{J}\mathbf{Z}_{f})))
\end{equation}
where the set $\{\mathbf{Z}_{f}\}$ is a sequence matrix, $f=0,1,\cdots,\infty$.  $\mathbf{I}$ is identity matrix. With the initial approximation $\mathbf{Z}_{0}$, it can converge to the Moore-Penrose inverse $\mathbf{J}^{+}$ in the third-order, satisfying $\Vert\mathbf{J}\mathbf{J}^{+}-\mathbf{J}\mathbf{Z}_{0}\Vert_{F}<1$.}
\end{theorem}

The proof of the \textbf{theorem}~\ref{th:4.1}, please see our supplementary material. By setting $\mathbf{Z}_{0} = \mathbf{J}^{\top}/(\Vert\mathbf{J}\Vert_1\Vert\mathbf{J}\Vert_\infty)$, it can ensures  that $\Vert\mathbf{I}-\mathbf{J}\mathbf{Z}_{0}\Vert_{F}<1$. When $\mathbf{J}$ is is non-singular, $\Vert\mathbf{J}\mathbf{J}^{+}-\mathbf{J}\mathbf{Z}_{0}\Vert_{F} = \Vert\mathbf{I}-\mathbf{J}\mathbf{Z}_{0}\Vert_{F}<1$. Let $\mathbf{J}^{+}$ be approximated by $\tilde{\mathbf{Z}}$ with~\cref{eq: 20}, our Newton-Raphson's iterative equation can be written as:
\begin{equation}
\label{eq:21}
{\xi}_{z+1} = {\xi}_{z}-\tilde{\mathbf{Z}}({\xi}_{z})\mathcal{F}({\xi}_{z})
\end{equation}

According to the~\cref{eq:21}, we can implement the GPU-friendly Newton-Raphson method, where the input-to-output mapping is defined as: $\xi_{0} \in \mathbb{R}^{C \times 1 \times 6}\rightarrow \xi_{{final}} \in \mathbb{R}^{C \times 1 \times 6}$, where $C$ denotes the batch dimension for parallel solving.
 The final ultra-high precision pose $\mathbf{T}_{{final}}$ can be obtained through $\mathbf{T}_{{final}} = \text{exp}({\xi^{\wedge}_{{final}}}) $. 

For a convenient understanding of readers, we provide the overall algorithm flow in supplementary material. 
\section{Experiments}
\subsection{Dataset}
We provided a large-scale Gough-Stewart forward kinematics dataset for model training. In order to obtain corresponding relational data, inverse kinematics problem in this dataset was solved by providing randomly limited motion space end-effector poses. To build the graph distance matrix, additional information about known configuration lengths was supplied. Where translation $\mathbf{t} \sim \mathcal{U}(l_{\text{min}}, l_{\text{max}})$ and rotation (Euler-angles representation) $\mathbf{q}_{3D} \sim \mathcal{U}(\theta_{\text{min}}, \theta_{\text{max}})$. In this work, we set $l_{\text{min}}=-50$mm, $l_{\text{max}}=50$mm, $\theta_{\text{min}} = -30$deg, and $\theta_{\text{max}} = 30$deg. For more details, please see the supplementary material. We have produced a large-scale dataset with a supplied size of $50\times10^{4}$ by applying the above-described process. The training set and test set have a ratio of $8:2$, with $40\times10^{4}$ and $10\times10^{4}$, respectively.

\subsection{Implement Details}
Our two-stage method is implemented using PyTorch framework. In DisGNet, we set the number of dense layers in the initialization block to $M=3$ and $M=2$ for the prediction head. For the message passing block, we employ the 2-KFW algorithm, conducting one 2-KFW algorithm within the block, with a feature dimension of $H=96$. We set the learning rate of the training model to 0.001, a scale factor $\beta = 250$ , and choose the Adam optimizer. We train our model for 1200 epochs with a batch size of 4000. In the Newton-Raphson method, the maximum number of iterations is set to $z_{\text{max}} = 100$, and the maximum iterations for $f_{\text{max}}=20$. Our experiments are conducted using a single RTX 3090 GPU.

\subsection{Metrics}
For forward kinematics learning, three metrics are proposed to evaluate and compare all models: ${E}$-trans, ${E}$-rot and ${E}$-ik.
$E\text{-trans}$  is a performance metric that calculates the average Euclidean distance between the estimated and ground truth values in the test set to evaluate translation estimation performance.
\begin{equation}
E\text{-trans} = \frac{1}{W}\sum_{\mathbf{t}\in\mathbb{R}^{3}}^{W}{\Vert{\mathbf{t}-\mathbf{t}_{gt}}\Vert}_2 
\end{equation}

$E\text{-rot}$ is a performance metric that calculates the average geodesic distance between the estimated and ground truth values in the test set to evaluate rotation estimation performance.
\begin{equation}
E\text{-rot} = \frac{1}{W}\sum_{\mathbf{R}\in{SO(3)}}^{W}{\Vert\text{log}({\mathbf{R}{\mathbf{R}^{\top}_{gt}}})\Vert}_{F}
\end{equation}

$E\text{-ik}$ is a performance metric that calculates the average Euclidean distance between the estimated joint variables and the ground truth joint variables in inverse kinematics to evaluate overall estimation performance.
\begin{equation}
E\text{-ik} = \frac{1}{W}\sum_{\mathbf{T}\in{SE(3)}}^{W}{\Vert(\mathcal{IK}(\mathbf{T})-\mathbf{l}_{os})\Vert}_{2}
\end{equation}
where $W$ represents the amount of samples in the test set.
\subsection{Experimental Results}
\begin{figure}[t]
\centering
\includegraphics[width=8.6cm]{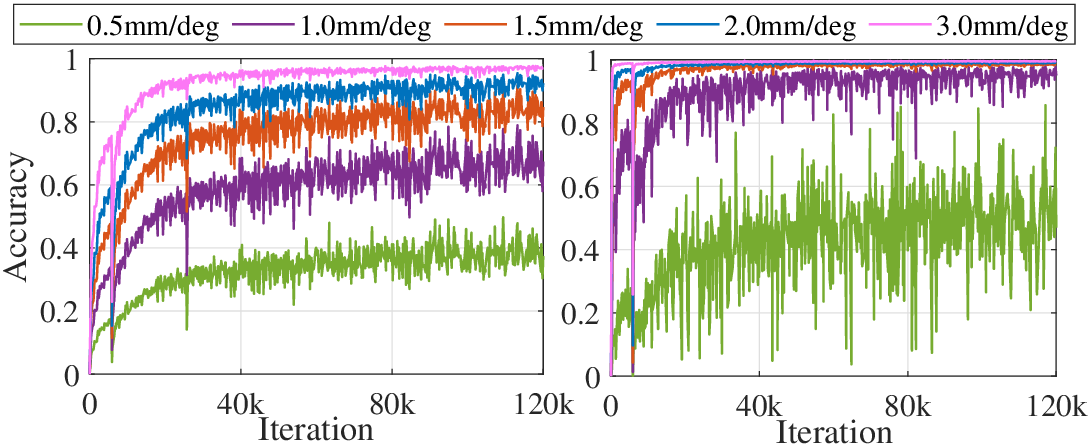}
\caption{The error accuracy of DisGNet on the test set during the training process. \textbf{Left:} Accuracy of translation errors (absolute error). \textbf{Right:} Accuracy of rotation errors (geodesic distance).} 
\label{fig:5}
\end{figure}

\textbf{Error accuracy.} In this paper, error accuracy refers to the percentage of samples in the overall test set with errors less than a certain threshold. In~\cref{fig:5}, we present the error accuracy under different thresholds: 0.5mm/deg, 1mm/deg, 1.5mm/deg, 2mm/deg, 3mm/deg. Specifically, DisGNet ensures that the percentage of samples with translation errors less than 1mm is 79.8\%, and the percentage of samples with rotation errors less than 1deg is 98.2\%.
\begin{figure}[H]
\centering
\includegraphics[width=8.7cm]{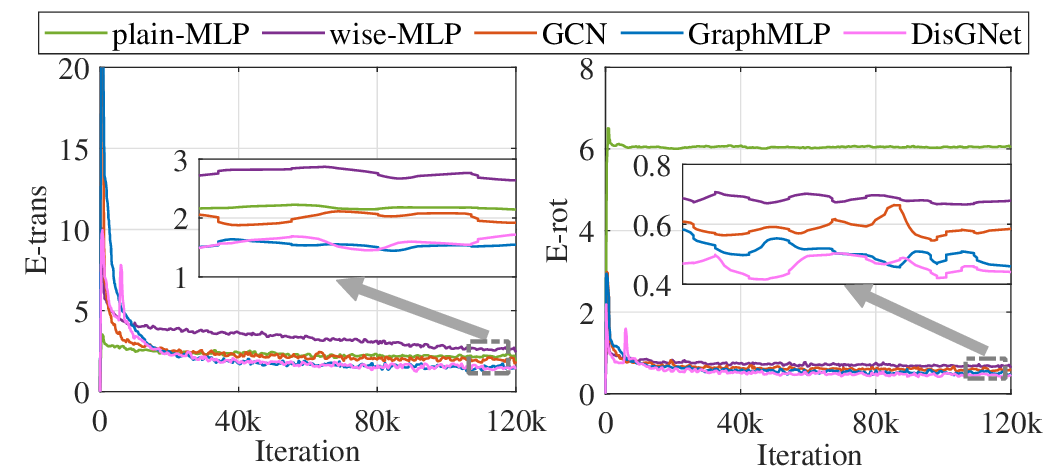}
\caption{The $E$-tans and $E$-rot of different networks on the test set during the training process. \textbf{Left:} The results of $E$-trans. \textbf{Right:} The results of $E$-rot.} 
\label{fig:6}
\end{figure}

\textbf{Comparison of different networks.} To highlight the superior performance of DisGNet under similar parameters, we compared three networks: plain-MLP, wise-MLP, and GCN~\cite{kipf2016semi}. The distinction between plain-MLP and wise-MLP lies in the fact that plain-MLP~\cite{seng1997forward} is a commonly used method in forward kinematics learning. It employs Euler angles as the rotation representation and utilizes MSE for learning. Its structure is defined as $\mathbb{R}^6\rightarrow(MLP+BN+GELU)\rightarrow\mathbb{R}^6$. A wise-MLP incorporates more advanced learning techniques, including the addition of residual connection, layer normalization, the use of a 9D rotation representation, and computation with the weighted loss function proposed by us. Its structure is similar to the dense layer in DisGNet. Moreover, we utilize GraphMLP~\cite{li2022graphmlp} to demonstrate that DisGNet can approximate results comparable to networks with larger parameter sizes and long-range learning capabilities.
\begin{table}[t]
\centering
\begin{threeparttable} 
\begin{tabular}{@{}cccccccc@{}}
\toprule
Method&Params&$E$-trans$\downarrow$&$E$-rot$\downarrow$&$E$-ik$\downarrow$&Time [s]\\
\midrule
plain-MLP&0.13M&2.03&6.13&36.50&0.0217\\
wise-MLP&0.13M&2.41&0.64&1.72&0.0293\\
GCN&0.13M&1.82&0.57&1.48&0.0525\\ 
GraphMLP&0.35M&\textbf{1.43}&0.44&1.19&0.1026\\ 
Ours (DisGNet)&0.13M& {1.45}&\textbf{0.41}&\textbf{1.14}&0.0814\\
\bottomrule
\end{tabular}
\caption{Quantitative results for various networks. Million parameters are denoted as M.}
\label{tab:1}
\end{threeparttable} 
\end{table}
In~\cref{fig:6}, we present a comparison of two metrics on the test set for different networks, namely $E$-trans and $E$-rot. Through this comparison, we observe that DisGNet can maintain results comparable to the larger-parameter network GraphMLP, and outperform networks with similar parameters. Additionally, by using a 9D rotation representation, MLP also demonstrates the capability for rotation regression, significantly surpassing the rotation learning ability of plain-MLP. \cref{tab:1} shows quantitative results for a more exact comparison. In besides the three metrics, we included a comparison of parameters and network runtime. In comparison with GCN, DisGNet outperforms it in terms of distance information learning. Compared to plain-MLP, DisGNet achieves a reduction of 28.5\% in $E$-trans, 93.3\% in $E$-rot, and 96.8\% in $E$-ik.
\begin{figure}[H]
\centering
\includegraphics[width=8.6cm]{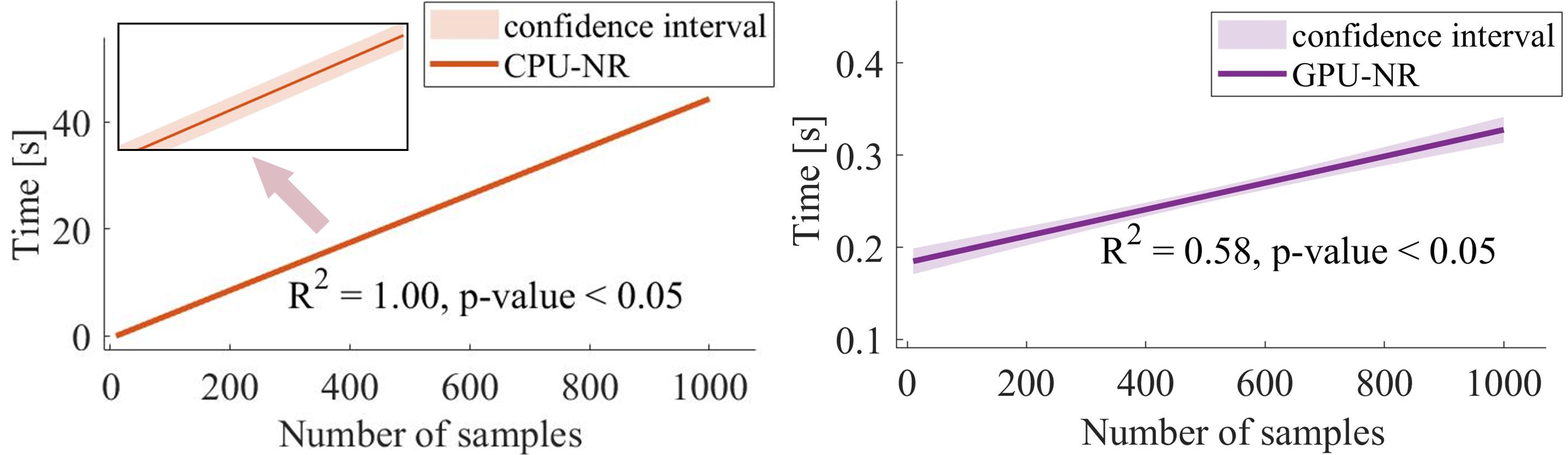}
\caption{The time consumption comparison of the Newton-Raphson method on different computing platforms for various numbers of examples. \textbf{Left:} Traditional Newton-Raphson method (performed on the CPU platform). \textbf{Right:} GPU-friendly Newton-Raphson method (performed on the GPU platform). Precision level  $\gamma = 0.001$ and initialize values are given by DisGNet.} 
\label{fig:7}
\end{figure}

\textbf{Comparison of different Newton-Rapson methods.}  \cref{fig:7} shows the Newton-Raphson runtimes for two types of methods: Traditional Newton-Raphson (CPU-NR) and GPU-friendly Newton-Raphson (GPU-NR)  for various number of samples on different computing platforms. The CPU-NR runtime increases with the number of samples, while the GPU-NR runtime does not increase as quickly, guaranteeing that the calculation can be finished in 0.4 seconds. This proves the effectiveness of CUDA's parallel computing when dealing with large-scale data as input. Please note that the GPU runtime is not linear. What we have provided is the trend after linear regression with a 95\% confidence interval.
\begin{table}[H]
\centering
\begin{threeparttable} 
\begin{tabular}{@{}cccccccc@{}}
\toprule
Method&P-level&Average time [ms]&Success rate\\
\midrule
MLP+CPU-NR~\cite{parikh2005hybrid}&$\gamma = 0.0001$&54.83&82.1\%\\
DisGNet+GPU-NR&$\gamma = 0.0001$& {0.51}&96.5\%\\
\bottomrule
\end{tabular}
\caption{The results of different two-stage methods. P-level refers to the the precision level.}
\label{tab:2}
\end{threeparttable} 
\end{table}
\textbf{Comparison of different two-stage methods.} \cref{tab:2} provides a comparison with the traditional two-stage method, including average time and success rate. We set the precision level 
$\gamma$ to 0.0001 and calculate it for 1000 samples. In the case of parallel computation for 1000 samples without encountering singular Jacobian matrices, our method only requires an average time of 0.51 ms. In the case of non-parallel computation for 1000 samples with singular Jacobian matrices, our method can guarantee a success rate of 96.5\%.
\section{Conclusion}
For the forward kinematics problem of parallel manipulator, we construct the graph for the parallel manipulator using complex structure and use its known graph distance matrix as the learning feature. To effectively learn the distance matrix, we propose DisGNet, which utilizes the 
k-FWL test for message-passing to obtain higher expressive power for achieving high-precision pose output. To meet the real-time requirements, we extend the Newton-Raphson method to the GPU platform, enabling it to work in conjunction with DisGNet and obtain ultra-high precision poses.


\addtolength{\textheight}{-12cm}   







\bibliographystyle{IEEEtran}
\bibliography{mybib}

\end{document}


\maketitle
\thispagestyle{empty}
\pagestyle{empty}

\begin{appendix}
\subsection{Proof}

\newtheorem{theorem}{\bf  Theorem}

\begin{theorem}
\label{theorem:1}
\textit{For Jacobian matrix $\mathbf{J}\in\mathbb{R}^{6\times6}$, given the initial approximation ${\mathbf{Z}_{0}}$,  we have:
\begin{equation}
\label{eq: 20}
 \mathbf{Z}_{f+1}=\frac{1}{4}\mathbf{Z}_{f}(13\mathbf{I}-\mathbf{J}\mathbf{Z}_{f}(15\mathbf{I}-\mathbf{J}\mathbf{Z}_{f}(7\mathbf{I}-\mathbf{J}\mathbf{Z}_{f})))
\end{equation}
where the set $\{\mathbf{Z}_{f}\}$ is a sequence matrix, $f=0,1,\cdots,\infty$.  $\mathbf{I}$ is identity matrix. With the initial approximation $\mathbf{Z}_{0}$, it can converge to the Moore-Penrose inverse $\mathbf{J}^{+}$ in the third-order, satisfying $\Vert\mathbf{J}\mathbf{J}^{+}-\mathbf{J}\mathbf{Z}_{0}\Vert_{F}<1$.}
\end{theorem}
\textit{Poof.} For proving the convergence of~\cref{eq: 20}, consider first that $\Vert\mathbf{I}-\mathbf{J}\mathbf{Z}_0\Vert_F<1$, $\mathbf{H}_{0} = \mathbf{I}-\mathbf{J}\mathbf{Z}_0$, $\mathbf{H}_{f} = \mathbf{I}-\mathbf{J}\mathbf{Z}_f$. For~\cref{eq: 20}, we can have:
\begin{align}
&\mathbf{H}_{f+1} = \mathbf{I}-\mathbf{J}\mathbf{Z}_{f+1}\notag\\
&=\mathbf{I}-\mathbf{J}(\frac{1}{4}\mathbf{Z}_{f}(13\mathbf{I}-\mathbf{J}\mathbf{Z}_{f}(15\mathbf{I}-\mathbf{J}\mathbf{Z}_{f}(7\mathbf{I}-\mathbf{J}\mathbf{Z}_{f}))))\notag\\
&=\mathbf{I}-\mathbf{J}(-\frac{1}{4}\mathbf{Z}_{f}(-13\mathbf{I}+15\mathbf{J}\mathbf{Z}_{f}-7(\mathbf{J}\mathbf{Z}_{f})^{2}+(\mathbf{J}\mathbf{Z}_{f})^{3}))\notag\\
&=\frac{1}{4}(4\mathbf{I}-\mathbf{J}\mathbf{Z}_f)(\mathbf{I}-\mathbf{J}\mathbf{Z}_f)^{3}\notag\\
&=\frac{1}{4}(3\mathbf{I}+\mathbf{I}-\mathbf{J}\mathbf{Z}_{f})(\mathbf{I}-\mathbf{J}\mathbf{Z}_f)^{3}\notag\\
&=\frac{1}{4}(3\mathbf{I}+\mathbf{H}_f)(\mathbf{H}_f)^{3}\notag\\
&=\frac{1}{4}(3\mathbf{H}^{3}_f+\mathbf{H}^{4}_f)
\label{eq:2}
\end{align}
thus, we get $\Vert\mathbf{H}_{f+1}\Vert = \frac{1}{4}(\Vert3\mathbf{H}^{3}_f+\mathbf{H}^{4}_f\Vert)\leq\frac{1}{4}(3\Vert\mathbf{H}^{3}_f\Vert+\Vert\mathbf{H}^{4}_f\Vert)$. And since $\Vert\mathbf{H}_0\Vert<1$, $\Vert\mathbf{H}_1\Vert\leq\Vert\mathbf{H}_0\Vert^{3}<1$, we can have:
\begin{equation}
\Vert\mathbf{H}_{f+1}\Vert\leq\Vert\mathbf{H}_f\Vert^{3}\leq\Vert\mathbf{H}_{f-1}\Vert^{3^{2}}\leq\cdots\leq\Vert\mathbf{H}_0\Vert^{3^{f+1}}<1
\label{eq:3}
\end{equation}
where $\mathop{\lim}\limits_{f\rightarrow\infty}\Vert\mathbf{H}_{f}\Vert=0$, \eg, $\mathop{\lim}\limits_{f\rightarrow\infty}\Vert\mathbf{I}-\mathbf{J}\mathbf{Z}_f\Vert=0$, $\mathop{\lim}\limits_{f\rightarrow\infty}\mathbf{I}-\mathbf{J}\mathbf{Z}_f=0$. Thus, for~\cref{eq: 20}, we can attain $\mathbf{Z}_f = \mathbf{J}^{+}$, when $f\rightarrow\infty$. The proof of convergence is complete.

Then, we proving the convergence is third-order. Firstly, we define $\epsilon_f = \mathbf{J}^{+} - \mathbf{Z}_f$ is error matrix in~\cref{eq: 20}. Apply the same procedure as in~\cref{eq:2} we can have:
\begin{equation}
\mathbf{I}-\mathbf{J}\mathbf{Z}_{f+1} = \frac{1}{4}(3(\mathbf{I}-\mathbf{J}\mathbf{Z}_{f})^{3}+(\mathbf{I}-\mathbf{J}\mathbf{Z}_{f})^{4})
\end{equation}
now, we can obtain $\mathbf{J}\epsilon_{f+1}$:
\begin{align}
&\mathbf{J}\epsilon_{f+1}= \mathbf{J}(\mathbf{J}^{+}-\mathbf{Z}_{f+1})\notag\\
&=\frac{1}{4}(3(\mathbf{J}(\mathbf{J}^{+}-\mathbf{Z}_{f}))^{3}+(\mathbf{J}(\mathbf{J}^{+}-\mathbf{Z}_{f}))^{4})\notag\\
&=\frac{1}{4}(3(\mathbf{J}\epsilon_f)^3+(\mathbf{J}\epsilon_f)^{4})
\label{eq:5}
\end{align}
according to~\cref{eq:5}, we can get $\epsilon_{f+1} $:
\begin{equation}
\epsilon_{f+1}= \frac{1}{4}(3\epsilon_f(\mathbf{J}\epsilon_f)^2+\epsilon_f(\mathbf{J}\epsilon_f)^3)
\end{equation}
where $\Vert\epsilon_{f+1} \Vert\leq\frac{1}{4}(3\Vert\epsilon_f\Vert\cdot\Vert\mathbf{J}\epsilon_f\Vert^2+\Vert\epsilon_f\Vert\cdot\Vert\mathbf{J}\epsilon_f\Vert^3) $, and subsequently:
\begin{equation}
\Vert\epsilon_{f+1} \Vert\leq(\frac{1}{4}(3\Vert\mathbf{J}\Vert^2+\Vert\mathbf{J}\Vert^3\Vert\epsilon_f\Vert))\Vert\epsilon_f\Vert^3
\end{equation}
this indicates that is at least a third order of convergence for the iterative~\cref{eq: 20} as it approximates to $\mathbf{J}^{+}$. This completes the proof.

\subsection{Algorithm flow}
We provide the process of our two-stage solver algorithm and the randomized generation dataset technique below to help readers better grasp our overall algorithm and dataset.
\begin{algorithm}
\floatname{algorithm}{Algorithm}
\renewcommand{\algorithmicrequire}{\textbf{Input:}}
\renewcommand{\algorithmicensure}{\textbf{Otput:}}
\footnotesize
\caption{Two-stage solver }
\label{alg1}
\begin{algorithmic}[1]
    \REQUIRE Graph distance matrix $\mathbf{E}$;
    \ENSURE Ultra-high precision pose of the end-effector $\mathbf{T}_{final}$;
    \STATE $\mathbf{T}=\mathcal{G}(f_{init}(\mathbf{E}))$; 
	\STATE $f_\text{max}=20$, $z_\text{max}=100$, $\text{batch size}=C$,  $\text{precision level} = \gamma$; 
    \STATE $\xi = \text{log}(\mathbf{T})^{\vee}$; 
 \FOR{each $z \leq z_\text{max}$}  
\STATE $\mathbf{J} = \mathbf{J}(\xi)$; $\rhd$ Get Jacobian matrix
	\FOR{each $f \leq f_\text{max}$} 
	\STATE $\mathbf{Z}_{f+1}=\frac{1}{4}\mathbf{Z}_{f}(13\mathbf{I}-\mathbf{J}\mathbf{Z}_{f}(15\mathbf{I}-\mathbf{J}\mathbf{Z}_{f}(7\mathbf{I}-\mathbf{J}\mathbf{Z}_{f})))$;
\IF{Average$(\Vert\mathbf{J}\mathbf{Z}_{f+1}-\mathbf{J}\mathbf{Z}_{0}\Vert_{F}<1)$} 
\STATE $\tilde{\mathbf{Z}} = \mathbf{Z}_{f+1}$;  
\ENDIF
	\ENDFOR  
\STATE ${\xi}_{z+1} = {\xi}_{z}-\tilde{\mathbf{Z}}({\xi}_{z})\mathcal{F}({\xi}_{z})$;
\IF{$({\Vert {\xi}_{z+1}-{\xi}_{z}\Vert_1}<\gamma)$} 
\STATE $\xi_{{final}} = {\xi}_{z+1}$;  
\ENDIF
\ENDFOR  
    \RETURN $\mathbf{T}_{{final}} = \text{exp}(\xi^{\wedge}_{{final}})$;
\end{algorithmic}
\end{algorithm}
\begin{algorithm}
\floatname{algorithm}{Algorithm}
\renewcommand{\algorithmicrequire}{\textbf{Parameters:}}
\renewcommand{\algorithmicensure}{\textbf{Result:}}
\footnotesize
\caption{Randomized generation dataset}
\label{alg1}
\begin{algorithmic}[1]
    \REQUIRE $l_{\min},l_{\max},\theta_{\min},\theta_{\max}, \mathbf{\Omega}$;
    \ENSURE Dataset for the kinematics of Gough-Stewart platform;
    \STATE $\mathbf{x}\sim\mathcal{U}(l_{\min},l_{\max})$;  $\rhd$ Translation value $\mathbf{x} = (x\ y\ z)\in\mathbb{R}^{3}$;
    \STATE $\mathbf{q}_{3D}\sim\mathcal{U}(\theta_{\min},\theta_{\max})$; $\rhd$ Rotation value (Euler-angles representation)  $\mathbf{q}_{3D} =(\alpha\ \beta\ \gamma)\in\mathbb{R}^{3}$;
    \STATE $\mathbf{R}\leftarrow\mathbf{R}_{x}(\alpha)\mathbf{R}_{y}(\beta)\mathbf{R}_{z}(\gamma)\in SO(3)$; 
    \STATE $\mathbf{\bar{l}}_s, \mathbf{l}_{os}\leftarrow\mathcal{IK}(\mathbf{\Omega},\mathbf{x}, \mathbf{R})$ $\rhd$ Platform configuration information $\mathbf{\Omega}=(\mathbf{a}_s\ \mathbf{b}_s)$, Inverse kinematics solving function $\mathcal{IK}(\cdot)$; 
    \STATE $\mathbf{q}_{4D}\leftarrow\varPhi(\mathbf{R})$; $\rhd$ $\varPhi$ is Quaternion mapping, Quaternion $\mathbf{q}_{4D}\in\mathcal{S}^{3}$;
    \STATE $\mathbf{E} =\mathcal{P}(\mathcal{V}, \mathcal{E},d_{ij})$; $\rhd$ Distance matrix (adjacent matrix with edge-length information) $\mathbf{E}$;
    \STATE $\mathbf{e}= \text{Vec}(\mathbf{E})$; $\rhd$ Vectorization operator $\text{Vec}(\cdot)$;
    \RETURN $\text{Concat}(\mathbf{x}, \mathbf{q}_{4D}, \mathbf{\bar{l}}_s, \mathbf{e})$ or $\text{Concat}(\mathbf{x}, \mathbf{q}_{3D}, \mathbf{\bar{l}}_s, \mathbf{e})$; $\rhd$ Get a dataset for the kinematics of the Gough-Stewart platform;
\end{algorithmic}
\end{algorithm}

\begin{table*}[t]
\centering
\begin{threeparttable} 
\begin{tabular}{cccccccccccc}
\toprule
\multirow{2}{*}{Training Set}&\multirow{2}{*}{Method}&$\vert$&\multicolumn{4}{c}{{\underline{Translation Error (mm)}}}&$\vert$&\multicolumn{4}{c}{{\underline{Rotation Error (deg)}}} \\
&&$\vert$&Median$\downarrow$&Mean$\downarrow$&Acc(0.5mm)$\uparrow$&Acc(1mm)$\uparrow$&$\vert$&Median$\downarrow$&Mean$\downarrow$&Acc($\text{0.5}^{\circ}$)$\uparrow$&Acc($\text{1}^{\circ}$)$\uparrow$\\
\midrule
\multirow{5}{*}{Full}&plain-MLP&&0.72&1.01&33.2&61.4&&5.25&6.13&1.7&7.6\\
&wise-MLP&&0.89&1.32&26.7&54.6&&0.50&0.64&72.4&94.3\\
&GCN&&0.68&0.96&41.5&71.3&&0.42&0.57&76.4&95.1\\
&GraphMLP&&\textbf{0.48}&\textbf{0.73}&\textbf{52.4}&\textbf{80.5}&&\textbf{0.34}&{0.44}&\textbf{85.6}&98.0\\
&Ours (DisGNet)&&\textbf{0.48}&{0.75}&{51.7}&{79.8}&&{0.35}&\textbf{0.41}&{85.1}&\textbf{98.2}\\
\midrule
\multirow{5}{*}{90\%}&plain-MLP&&0.84&1.28&31.1&57.5&&6.08&7.12&0.9&5.4\\
&wise-MLP&&0.96&1.39&28.7&49.6&&0.58&0.88&47.4&91.7\\
&GCN&&0.78&1.08&34.3&60.1&&0.47&0.72&53.9&94.5\\
&GraphMLP&&\textbf{0.54}&\textbf{0.86}&45.1&\textbf{72.7}&&0.38&0.45&77.8&96.2\\
&Ours (DisGNet)&&{0.55}&{0.88}&\textbf{45.4}&{71.9}&&\textbf{0.36}&\textbf{0.44}&\textbf{78.6}&\textbf{97.2}\\
\midrule
\multirow{5}{*}{80\%}&plain-MLP&&0.97&1.51&25.7&53.9&&7.58&8.94&0.5&3.6\\
&wise-MLP&&1.25&2.38&20.1&42.5&&0.73&0.95&41.4&84.3\\
&GCN&&0.92&1.42&28.3&55.8&&0.64&0.82&43.6&88.5\\
&GraphMLP&&0.70&\textbf{0.95}&39.1&64.8&&0.45&0.55&55.3&94.2\\
&Ours (DisGNet)&&\textbf{0.69}&\textbf{0.95}&\textbf{39.3}&\textbf{65.1}&&\textbf{0.44}&\textbf{0.52}&\textbf{57.6}&\textbf{96.7}\\
\bottomrule
\end{tabular}
\caption{A comparison of methods by median, mean, 0.5mm and 1mm accuracy of errors, as well as $\text{0.5}^{\circ}$ and $\text{1}^{\circ}$ accuracy of (geodesic distance) errors. }
\label{tab:1}
\end{threeparttable} 
\end{table*}
\begin{figure*}[t]
\centering
\includegraphics[width=16.8cm]{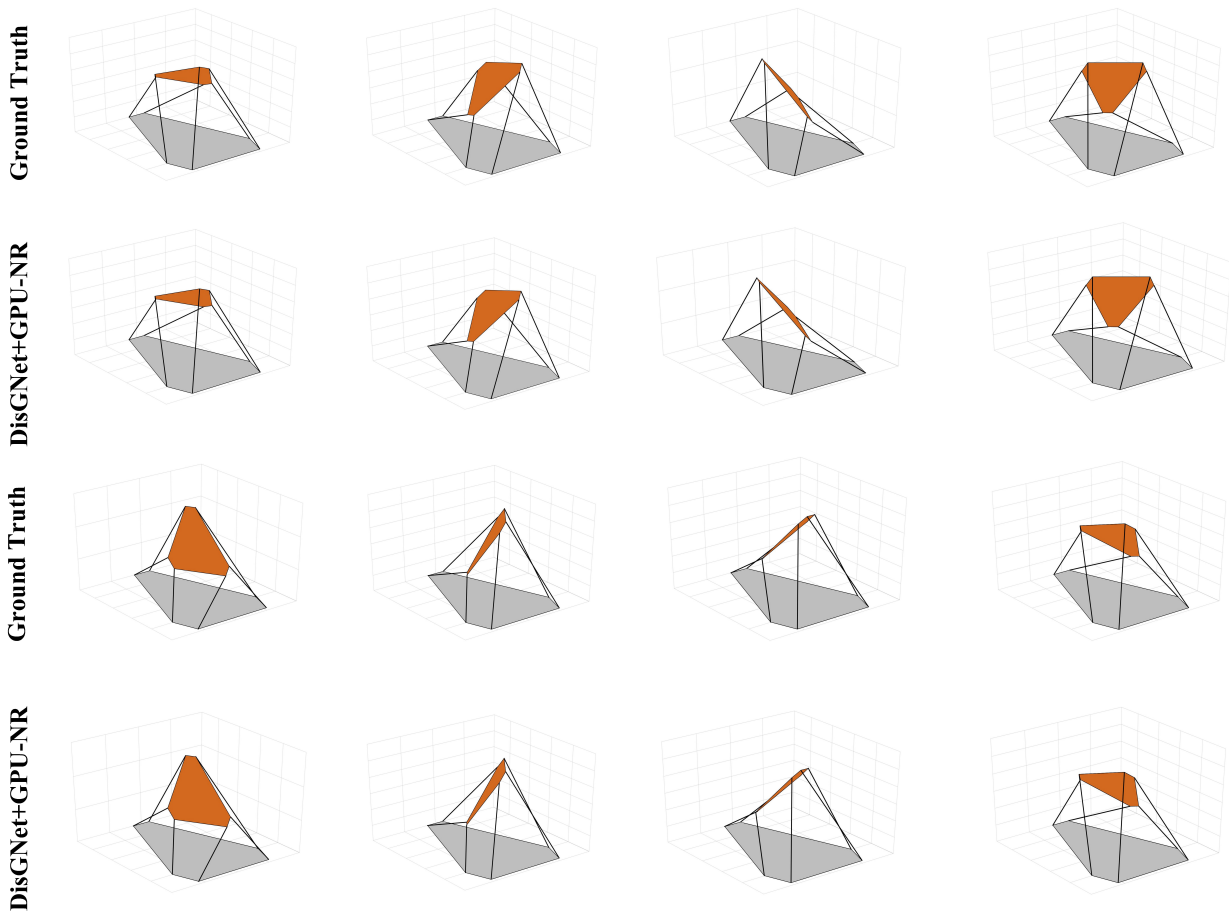}
\caption{Qualitative illustration and comparison for the pose of the GSP's end-effector. The results are obtained by our two-stage slover (DisGNet and GPU-friendlly Newton-Raphson) on the test set, the setting of precision level $\gamma$ is 0.0001.} 
\label{fig:1}
\end{figure*}
\begin{table*}[t]
\centering
\begin{threeparttable} 
\begin{tabular}{cccccccc}
\toprule
Method&Mapping (input-to-output)&w/ distance matrix?&Rotation representation&Loss function&w/ residuals?&Normalization\\
\midrule
plain-MLP&$\mathbb{R}^{6}\rightarrow\mathbb{R}^{6}$&w/o distance matrix &Euler angles&MSE&w/o residuals&BN\\
wise-MLP&$\mathbb{R}^{6}\rightarrow SE(3)$ &w/o distance matrix &9D-SVD&Weighted loss&w/ residuals&LN\\
GCN&$\mathbb{R}^{12\times12}\rightarrow SE(3)$&w/ distance matrix&9D-SVD&Weighted loss&w/ residuals&LN\\
GraphMLP&$\mathbb{R}^{12\times12}\rightarrow SE(3)$&w/ distance matrix&9D-SVD&Weighted loss&w/ residuals&LN\\
Ours (DisGNet)&$\mathbb{R}^{12\times12}\rightarrow SE(3)$&w/ distance matrix&9D-SVD&Weighted loss&w/ residuals&LN\\
\bottomrule
\end{tabular}
\caption{The technical comparison of different models in the experiments. BN represents batch normalization, and LN represents layer normalization.}
\label{tab:2}
\end{threeparttable} 
\end{table*}
\subsection{More results}

\textbf{Model results.} We provide test results for models trained on three different-sized datasets: full size, 90\%, and 80\%. \cref{tab:1} offers a comparative experiment for different models, including median, mean, 0.5mm and 1mm accuracy of errors, as well as $\text{0.5}^{\circ}$ and $\text{1}^{\circ}$ accuracy of (geodesic distance) errors. Across the three datasets, our DisGNet outperforms methods with similar parameters and achieves comparable performance to GraphMLP.

\textbf{Qualitative results.} In~\cref{fig:1}, we present qualitative results for 8 different poses obtained through DisNet+GPU-friendly Newton-Raphson. Here, the precision level is set to 0.0001. Our two-stage method can ensures that the overall pose aligns with the ground truth.

\begin{figure}[H]
\centering
\includegraphics[width=8.6cm]{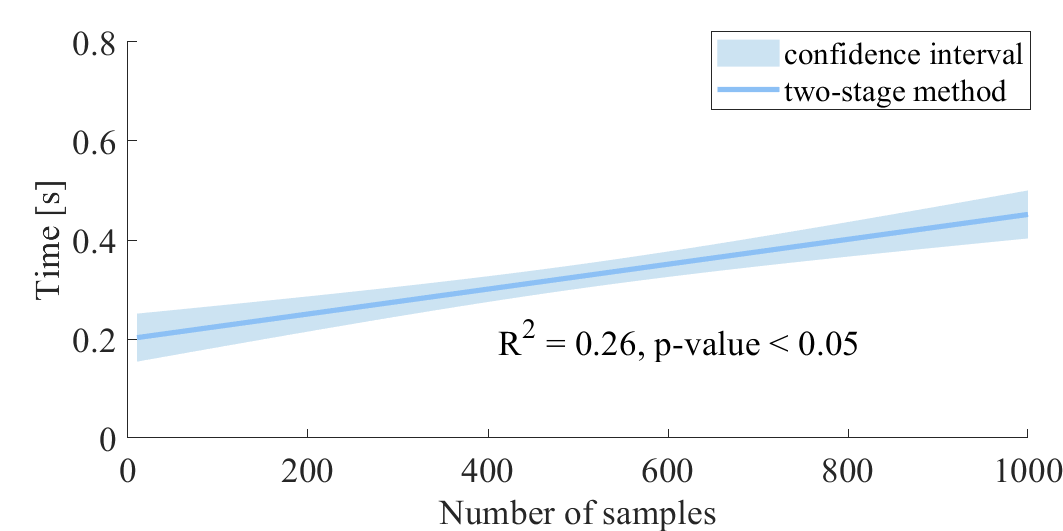}
\caption{The time taken by our two-stage slover in the different number of samples of parallel solving. The results are obtained by our two-stage slover (DisGNet and GPU-friendlly Newton-Raphson) on the test set, the setting of precision level $\gamma$ is 0.001. The 95\% confidence interval is shown by the shaded region.} 
\label{fig:2}
\end{figure}
\textbf{Overall time taken results.} In~\cref{fig:2}, we present the time taken by our two-stage slover in the different number of samples of parallel solving. Samples range from 10 to 1000, increasing by 10 at each step. As the number of samples increases, the solving time will not increase rapidly, ensuring that the calculation can be completed within 1 second. But it's important to note that GPU parallel processing time does not exhibit a linear trend. Therefore, we performed linear regression on the data to assess the overall trend and plotted the 95\% confidence interval for reference. Specifically, the solving time for 10 samples is 0.1464s, and for 1000 samples, it is 0.3649s. 

\begin{table}[H]
\centering
\begin{threeparttable} 
\begin{tabular}{@{}cccccccc@{}}
\toprule
Method&NoI ($0.01$)&NoI ($0.001$)&NoI ($0.0001$)\\
\midrule
MLP+CPU-NR~\cite{parikh2005hybrid}&13376&17035&19554\\
DisGNet+GPU-NR&15& 45&75\\
\bottomrule
\end{tabular}
\caption{The results of the required number of iterations (NoI) for various two-stage methods to achieve different precision levels. The precision levels are set at 0.01, 0.001, and 0.0001, respectively.}
\label{tab:3}
\end{threeparttable} 
\end{table}

\textbf{Number of iterations results.} In ~\cref{tab:3}, we present the number of iterations (NoI) required by two different two-stage methods for 1000 sample sets. The GPU's parallel computing capability significantly reduces the number of iterations. For the precision level $\gamma =0.0001$, the traditional method requires 260 times more iterations than our method.

\subsection{More details on experiments}
\textbf{Details of different networks.}  In ~\cref{tab:3}, we provide technical details of different networks in the experiments, including the mapping relationship, the use of distance information, rotation representation, loss function, and use of residual connections and normalization techniques. Vital learning techniques include 9D rotation representation~\cite{levinson2020analysis}, residual connections~\cite{he2016deep}, and layer normalization~\cite{ba2016layer}. Under appropriate conditions, these advanced learning techniques can significantly improve network performance.

\textbf{Details of GPU-friendly Newton-Raphson method.} In the paper, we employ~\textbf{theorem}~\ref{theorem:1} to approximate the Moore-Penrose pseudoinverse, efficiently computed on GPU through matrix-matrix multiplication. Specifically, with 20 iterations, we achieve a good approximation of the Moore-Penrose pseudoinverse. In addition, the comparative experiments for different Newton-Raphson methods follow the same settings as illustrated in~\cref{fig:2}, \ie, the number of samples ranges from 10 to 1000, increasing by 10 at each step.

\textbf{Details of the evaluation metrics.} Here, it is worth noting that the additional evaluation metrics refer to the comparison of different two-stage methods in terms of average time and success rate at a precision level of 0.0001. For the convenience of readers, we provide the calculation formulas for the two evaluation metrics as follows:
\begin{equation}
\text{Average time} = \frac{T}{W}
\end{equation}
\begin{equation}
\text{Success rate} = (1-\frac{N_1 + N_2}{W})\times100\%
\end{equation}
where $W$ denotes the total number of test samples (in this work, $W=1000$), and $T$ denotes the total runtime for the total number of test samples $W$. The test sample numbers $N_1$ and $N_2$ represent the number of singular Jacobian matrix occurrences and samples that do not meet the required precision level, respectively.

\subsection{Time complexity of DisGNet}
Regarding the time complexity of DisGNet, it is primarily determined by the k-FWL algorithm used in the message-passing block. According to the description in~\cite{li2023is, maron2019provably}, the complexity of the k-FWL algorithm is $O(kn^{k+1})$. We set $k=2$, thus the time complexity of DisGNet is $O(n^{3})$.

\subsection{More discussion}
\textbf{Limitations.} (i) The $O(n^{3})$ time complexity of the 2-FWL algorithm is the price DisGNet pays for its better performance in learning distance matrices when compared to traditional GCN. A careful consideration of the number of layers built for the message-passing process is required due to its higher complexity. Employing it as a deep network would not guarantee real-time performance. (ii) The advantage of the GPU-friendly Newton-Raphson method in optimizing poses lies in its efficiency when handling large-scale data via parallel computing. However, its advantages are not apparent when optimizing a single pose. Additionally, when encountering singular Jacobian matrices, GPU parallel computing is not feasible, requiring the early identification and exclusion of singular points.

\textbf{Future work.} Our next work is to explore how to efficiently utilize the distance matrix for learning. Due to the relatively high complexity of the 2-FWL algorithm, we need to further investigate methods that can efficiently leverage distance information and reduce computational costs. This is essential to enable faster training and deployment on practical hardware for computation.
\bibliographystyle{IEEEtran}
\bibliography{mybib}



\addtolength{\textheight}{-12cm}   




\end{appendix}
